\documentclass[letterpaper]{article}
\usepackage[preprint]{aaai2027}
\usepackage[hyphens]{url}
\usepackage{graphicx}
\usepackage{natbib}
\usepackage{caption}
\usepackage{float}
\usepackage{booktabs}
\usepackage{algorithm}
\usepackage{algorithmic}
\usepackage{amsmath}
\title{Implicit Minor Detection from Multi-Turn Dialogue in High-Risk AI Interactions: A Constrained Self-Evolving LLM Harness}
\author{
Xiaohan~Zhang,
Ying~Zhao,
Ruilin~Guo,
Xuanyuan~Yu,\\
Kaibo~Huang,
Yukun~Wei,
Zhongliang~Yang\thanks{Corresponding authors.},
Linna~Zhou\textsuperscript{*}
}
\affiliations{
Beijing University of Posts and Telecommunications\\
\texttt{\{zhangxiaohan2005, zhaoying, guoruilin, yuxuanyuan\}@bupt.edu.cn}\\
\texttt{\{huangkaibo, weiyukun, yangzl, zhoulinna\}@bupt.edu.cn}
}

\begin{document}
\maketitle

\begin{abstract}
In high-risk text interactions, a system may need to estimate whether the current speaker is likely to be a minor before applying age-conditioned safeguards. Many text-based dialogue systems lack strong identity signals such as account information, verified identity, facial data, or device data, and must rely on weak cues that emerge across multiple turns. We formulate implicit minor detection under weak identity information as a high-risk user-attribute inference task and propose a constrained framework comprising a large language model, a Harness, and a Maker-Checker protocol. The framework controls when inference is invoked, requires structured dialogue-linked evidence, and subjects validation-selected Harness revisions to version-level review. On a controlled Chinese benchmark comprising 8,328 multi-turn dialogues, our method improves average precision and minor-class F1 by 7.60 and 6.37 percentage points, respectively, over direct inference using the same LLM family.
\end{abstract}

\section{Introduction}

Generative AI systems for companionship, learning support, and mental-health assistance increasingly operate in settings where minors may participate. Applying suitable safeguards often requires an initial judgment about whether the current speaker is likely to be a minor. Platform-level age assurance can use account, identity, facial, or parental-control signals \cite{livingstone2024ageassurance}. Text-based services embedded in external applications often have access only to the current conversation. In these settings, life-stage information appears gradually through references to school schedules, parental restrictions, examinations, work, housing, and other contextual details.

\begin{figure}[t]
\centering
\includegraphics[width=\columnwidth]{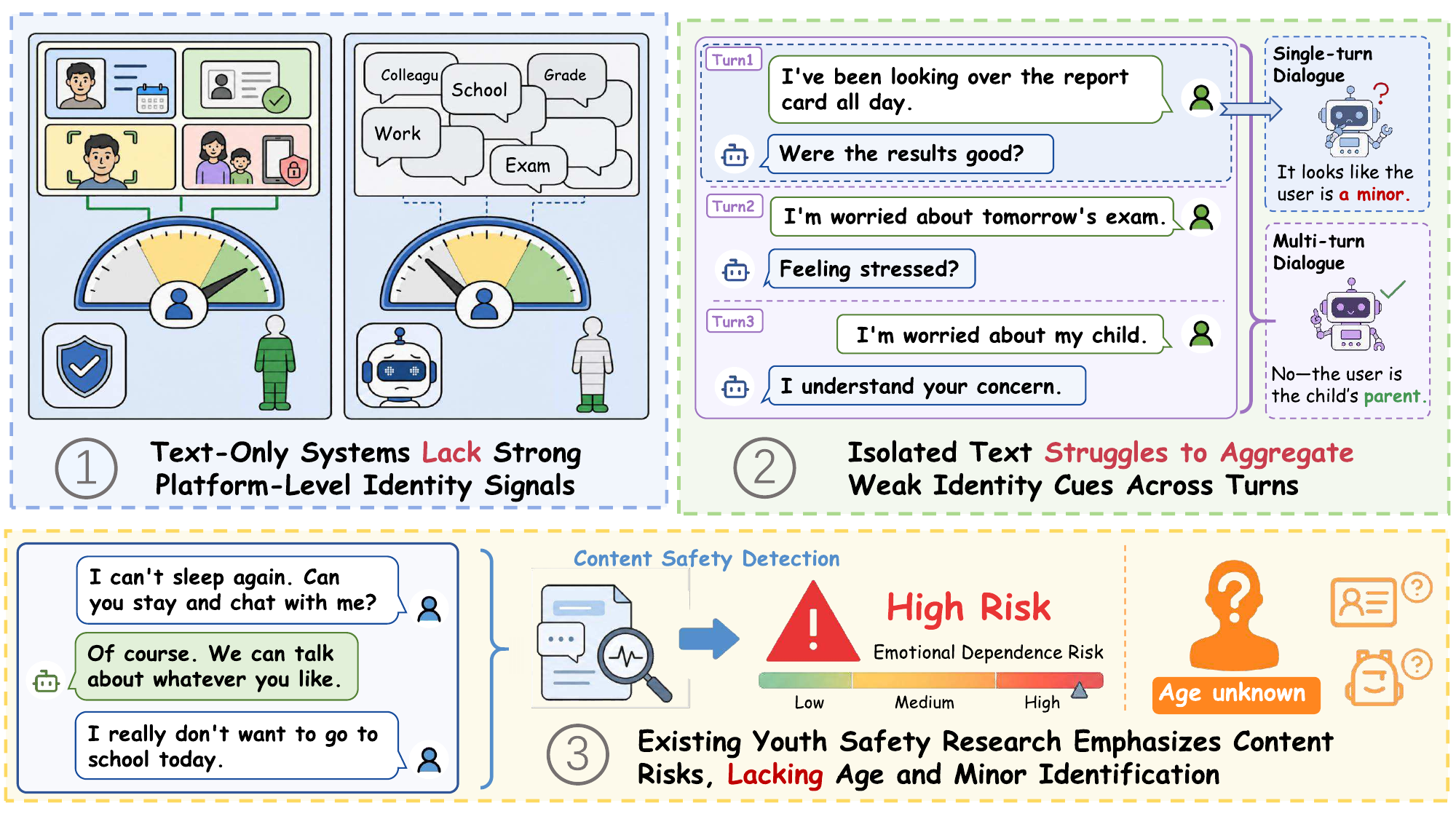}
\caption{Adjacent approaches leave a weak-identity gap. Platform age assurance depends on strong identity signals; static single-turn profiling loses cross-turn and speaker-binding context; content-safety safeguards assess content risk rather than the speaker's likely age.}
\label{fig:problem}
\end{figure}

Prior work covers three adjacent problems. Platform age assurance relies on account, identity, biometric, or parental-control signals \cite{livingstone2024ageassurance}. Textual author profiling infers demographic attributes from linguistic patterns, but usually treats the available text as a static profile \cite{rosso2017authorprofiling}. Youth-safety and content-safety research instead focuses on harmful content, moderation, and safeguards \cite{yu2025safeguarding,yu2025yair,yu2025youthsafe,han2024wildguard,inan2023llamaguard}. Figure~\ref{fig:problem} locates the gap between these lines of work.

Implicit minor detection poses four technical difficulties. First, individually weak cues may become informative only when combined across turns. Second, the system must bind each cue to the current speaker rather than to a parent, child, quoted source, or past self. Third, age inference should be invoked only when the request and dialogue justify it; a broad trigger risks unnecessary profiling, while a narrow trigger misses cases that need protection. Fourth, revisions are prone to regression: an edit that fixes one failure type can damage previously correct or boundary-sensitive cases.

We address these difficulties with a fixed LLM, a versioned Harness, and a Maker--Checker protocol. The Harness controls invocation, contextual support, evidence representation, and output validation. Its task-layer resources evolve from validation errors and recent revision history, while the promotion gate protects required behavior and rejects candidates that violate the admission constraints. Maker--Checker materials record version diffs and support inspection before the selected Harness is frozen.

To evaluate the task and framework, we construct a main benchmark and a separate trigger-boundary set. The main benchmark measures speaker-level recognition and component effects; the trigger set measures when minor detection should be invoked. Relative to direct inference with the same LLM family, our method improves average precision (AP) by 7.60 percentage points and minor-class F1 by 6.37 points. Trigger evolution raises trigger F1 by 7.23 points. A 50-participant human challenge-set study provides complementary evidence about item difficulty and annotator disagreement.

Our contributions are fourfold. First, we introduce a constrained self-evolving framework composed of a fixed LLM, a versioned Harness, and a Maker--Checker protocol. Second, we formulate implicit minor detection as a speaker-bound, multi-turn inference task distinct from static profiling, content moderation, and platform age assurance. Third, we construct an 8,328-dialogue benchmark and a separate 480-example trigger-boundary set for evaluating recognition and invocation. Fourth, experiments on recognition, trigger routing, component ablations, and version evolution show consistent gains over direct inference with the same LLM family; the human challenge set further identifies difficult and disputed cases.

\section{Related Work}

\paragraph{Dialogue attribute inference.}
Dialogue contains personal information that may be stated directly or implied across several utterances. Author-profiling benchmarks traditionally infer age, gender, or language variety from stable textual patterns \cite{rosso2017authorprofiling}. Conversation-focused methods infer open-ended personal facts or extract persona attributes from interaction histories \cite{tigunova2020charm,wang2022personalattributes,zhu2023paed}. Dialogue representations can also reveal a speaker's private persona \cite{li2022privatepersona}, while modern LLMs can recover sensitive properties from indirect cues \cite{staab2024privacy}. Minor status adds ambiguity because apparent evidence may be hypothetical, retrospective, or attributed to another person. Evidence-grounded NLP benchmarks evaluate predictions together with their supporting text \cite{deyoung2020eraser}, and MT-Bench-101 shows that turn-aware analysis can reveal behavior obscured by aggregate dialogue scores \cite{bai2024mtbench101}. We formulate minor detection as an evidence-grounded dialogue-level decision and evaluate difficult, adversarial, and boundary cases.

\paragraph{Age assurance and youth safety.}
Age-assurance research studies how platforms combine verification technologies, governance, children's rights, privacy, and public acceptance \cite{livingstone2024ageassurance,jarvie2024ageverification}. Youth-focused generative-AI research instead characterizes risks and parental controls \cite{yu2025safeguarding,yu2025yair}; YouthSafe advances this line with a benchmark and safeguard model \cite{yu2025youthsafe}. Their central question is whether content or model behavior is safe for youth. Our setting studies an earlier decision in the policy chain: whether age-conditioned protection is relevant when the speaker's age is implicit. A moderation model may identify harmful content without resolving who is speaking or whether a youth-specific policy should be activated.

Content-safety classifiers and safeguard models assess whether text or model responses are harmful \cite{han2024wildguard,inan2023llamaguard}. Their labels concern content risk rather than the identity of the speaker. XSTest shows that safety systems can overgeneralize from superficial cues \cite{rottger2024xstest}. In minor detection, the analogous error is assigning minor status to an adult because a dialogue contains school, family, or emotional vocabulary. We therefore include adult controls, hard negatives, adversarial cases, and explicit boundary examples.

\paragraph{Harnesses, self-refinement, and version governance.}
Existing methods improve individual responses or task programs through iterative critique, feedback memory, tool use, and metric-driven prompt optimization \cite{madaan2023selfrefine,shinn2023reflexion,yao2023react,khattab2023dspy,yuksekgonul2024textgrad,yang2024opro,gou2024critic}. Promptbreeder evolves prompt populations through mutation and selection \cite{fernando2024promptbreeder}. These methods establish the value of external feedback, but preserving previously correct behavior is not their central evaluation contract.

Our Harness treats optimization as a governed transition between external, versioned task resources. Candidate revisions are evaluated on frozen validation and regression cases, promoted only when predefined constraints are satisfied, and otherwise rejected with the incumbent retained. The next proposal receives both current errors and the recorded consequences of recent rejected or accepted edits. Behavioral testing motivates preserving targeted controls beyond aggregate accuracy \cite{ribeiro2020checklist}. Proposal, evaluation, and decision artifacts remain separate. Chain-of-Verification shows the value of an explicit verification stage for generated answers \cite{dhuliawala2024chainverification}; our framework applies verification to source-linked evidence at inference time and to candidate versions during development.

\section{Method}
\label{sec:method}

\subsection{Overall Framework}

We operationalize minor status as age below 18; speakers aged 18 or older belong to the adult class. Given a multi-turn dialogue window, the system has no access to account verification, facial signals, device metadata, or a trusted age attribute. We distinguish two decisions: whether minor detection should be invoked and, once invoked, whether the current speaker is likely to be a minor. Recognition also returns a confidence score and dialogue-grounded evidence.

Let $x$ denote a dialogue window. The task first predicts whether minor-status inference should be invoked, $g_{d_t}(x)\in\{0,1\}$, where $d_t$ is the current routing description. When $g_{d_t}(x)=1$, recognition returns a speaker label $\hat{y}\in\{0,1\}$, confidence $s\in[0,1]$, and dialogue-grounded evidence $e$, where $\hat{y}=1$ denotes minor and $s$ is the returned minor-confidence score. Our implementation additionally records a supported profile, ICBO fields, a conflict-resolution policy, and a follow-up action under the Harness contract. We write the system at iteration $t$ as
\begin{equation}
S_t=(L,H_t,\mathcal{M}),
\end{equation}
where $L$ is the fixed base LLM, $H_t$ is the versioned task Harness, and $\mathcal{M}$ denotes the Maker--Checker review protocol. Only $H_t$ is updated; the parameters of $L$ remain fixed.

\begin{figure*}[t]
\centering
\includegraphics[width=\textwidth]{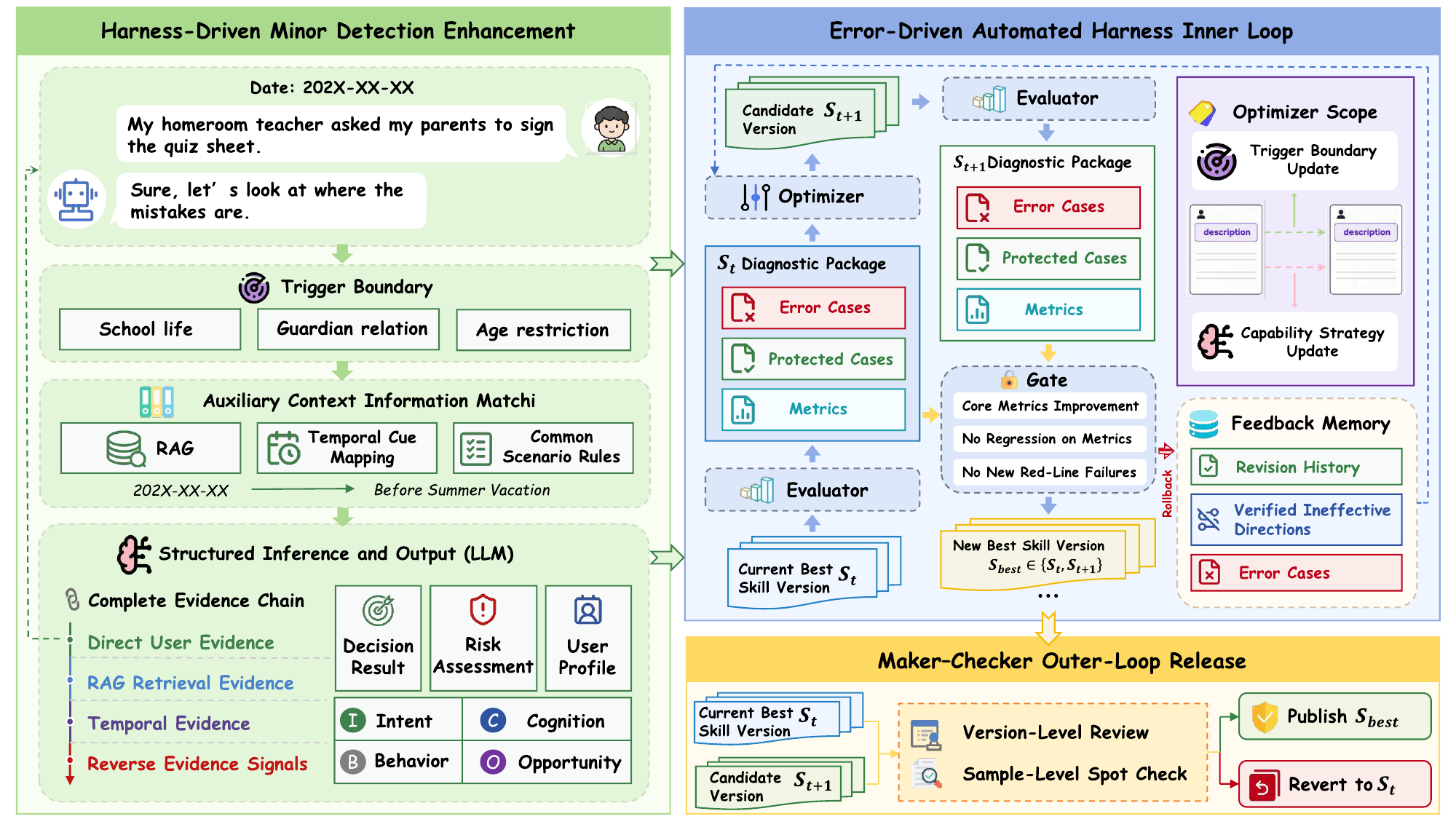}
\caption{Overview of the constrained implicit-minor-detection framework. Runtime inference separates trigger routing, contextual support, structured evidence, and source-constrained decisions. During development, validation errors and recent candidate outcomes guide restricted Harness revisions; the gate either promotes a candidate or retains the incumbent, and the selected version is accompanied by Maker--Checker review materials.}
\label{fig:framework}
\end{figure*}

Figure~\ref{fig:framework} summarizes inference and development. Each Harness version is evaluated through the same execution pipeline. The pipeline normalizes the input, derives optional temporal features, retrieves contextual cases, assembles an evidence package, invokes the classifier, validates the returned JSON, and repairs malformed output when necessary. This path remains fixed across versions, so observed differences come from changes to the Harness resources.

\subsection{Harness and Maker--Checker Protocol}

The Harness converts an open-ended model call into a constrained inference procedure with five components:
\begin{equation}
H_t=(\tau_t,C_t,\Phi_t,E_t,G_t),
\end{equation}
where $\tau_t$ controls the invocation boundary; $C_t$ organizes auxiliary context such as temporal features, task rules, and similarity retrieval; $\Phi_t$ defines the structured input--output contract; $E_t$ constrains decisive claims to traceable evidence; and $G_t$ proposes revisions, checks them for regression, and either promotes the candidate or retains the incumbent. The first four components govern a single inference. The final component governs transitions between Harness versions.

Contextual support combines temporal normalization, task rules, and similarity retrieval. Retrieval-augmented generation provides a standard mechanism for conditioning an LLM on selected external records \cite{lewis2020rag}. Our retriever uses \texttt{text-embedding-3-small} to select up to three cases from a case base built only from the 5,830-example training split. Retrieved labels, profiles, and utterances cannot be copied into the current user's profile or direct evidence. Retrieval cannot establish an age direction when the dialogue is insufficient or reverse a direction supported by current-dialogue evidence. We use it only for consistency checking.

The Maker--Checker protocol is separate from runtime inference. After validation-based selection, the implementation generates a version-diff review and an eight-item checklist that prioritizes changed predictions, low-confidence cases, and boundary-sensitive cases. The checklist covers label credibility, practical acceptability, boundary drift, explanation risk, the evaluation record, and the edited prompt text. These materials support manual version-level inspection; they do not alter runtime predictions or test metrics.

\subsection{Trigger Boundary}

School-related words, emotional language, or discussion of young people should not by themselves trigger minor detection. Over-routing would apply demographic inference to inputs for which it was neither requested nor justified. A description that requires an explicit age request would also miss legitimate implicit speaker-profiling tasks. We therefore model routing as a separate binary problem.

The routing description $d_t$ defines when the calling agent invokes $g_{d_t}$. A positive route requires that the request concern the age tendency, student stage, or minor-related risk of the current speaker or another explicitly designated dialogue participant. Topic-only educational discussion and incidental mentions of third parties are negative. Several mutually consistent weak signals may justify invocation, while a single school term, emotional expression, or family reference does not. The recognition benchmark itself remains current-speaker-only.

We optimize this boundary independently from downstream recognition. Description evolution changes only $d_t$ and is evaluated on a dedicated trigger dataset. Each example contains a user request, a gold invocation decision, and a routing scenario. Evaluation records whether the agent invokes the skill, whether invocation completes, and whether the routing output satisfies its contract. The recognition pipeline remains fixed so routing and recognition errors can be analyzed separately.

\subsection{Structured Evidence and ICBO Representation}

After invocation, the system binds each candidate cue to its subject. Mentions of another person, retrospective school memories, quoted material, and assistant-generated speculation cannot support a claim about the current speaker. Strong current-user anchors, such as an explicit K--12 task or a stable adult responsibility, receive priority. Emotion, dependence, family conflict, communication style, and similar weak cues contribute only when several of them support a consistent account of the speaker's current circumstances and no stronger cue contradicts it.

ICBO is a task-specific scaffold for organizing this evidence. Intention records what the speaker is trying to accomplish. Cognition records self-descriptions and identity-relevant mental context supported by the dialogue. Behavior summarizes relevant activity patterns. Opportunity/time represents temporal and situational constraints, including school schedules, work shifts, examinations, and recurring session timing. ICBO is an explicit intermediate record, not a learned psychological representation.

The evidence record separates direct dialogue evidence, temporal evidence, historical context, retrieved cases, and conflicting signals. Every direct-evidence item must be a short quotation or near-verbatim span locatable in the current dialogue. Summaries and inferences occupy separate fields, preventing a retrieved case or generated explanation from being presented as a user statement. The user profile follows the same rule and leaves fields unspecified when no dialogue-grounded identity anchor exists.

The decision policy records minor cues, adult cues, evidence sufficiency, conflict resolution, and the final decision basis. It combines subject-bound, typed evidence with the ICBO record to produce the binary label and minor-confidence score.
If evidence is marked insufficient, the frozen binary protocol maps the output to the negative class. This means that the dialogue does not support minor status; it does not verify adulthood. The benchmark does not evaluate a separate abstention class. A deployed system could instead request clarification or apply a conservative policy. The classifier receives the evidence package and output schema in one constrained call and returns one JSON object. A validator checks required fields, types, enumerated values, confidence ranges, and cross-field consistency. Schema repair is invoked only when this contract is violated.

\subsection{Gated Harness Evolution}

We improve the Harness through error-driven, validation-only evolution. Let $J(H_t;D_{\mathrm{val}})$ be the current validation report and let $U(H_t)$ denote the admission metrics: F1, invocation success, and step compliance. The report also contains contract statistics and indexed error artifacts. At each round, DeepSeek-V3 receives four forms of feedback: the incumbent's current errors and their failure types; currently correct validation cases that the revision should preserve; new errors introduced by the preceding candidate; and the diffs, metric changes, and acceptance or rejection reasons from the three most recent rounds. Thus, a proposal is informed by both unresolved errors and the observed consequences of earlier edits. The optimizer does not modify the base-model weights, output schema, controller, or evaluation code.

Here, self-evolution refers to validation-guided revision of Harness resources; the base model and evaluation code remain fixed.

The inner loop combines utility comparison with an explicit admission gate. $G_{\mathrm{core+}}=1$ when at least one component of $U(H'_t)$ improves over $U(H_t)$. For the main recognition task, F1 must also remain non-decreasing, as must invocation success, step compliance, and schema validity; no new redline failure may be introduced:
\begin{equation}
\begin{aligned}
\operatorname{Promote}(H'_t,H_t) ={}& G_{\mathrm{core+}} \land G_{\mathrm{F1}} \land G_{\mathrm{invoke}} \\
&\land G_{\mathrm{step}} \land G_{\mathrm{schema}} \land G_{\mathrm{redline}}.
\end{aligned}
\end{equation}
Equivalently,
\begin{equation}
H_{t+1}=\begin{cases}
H'_t, & \operatorname{Promote}(H'_t,H_t)=1,\\
H_t, & \text{otherwise}.
\end{cases}
\end{equation}
Decision summaries and new regression cases are retained for later proposals. In the reported runs, the loop stops after the round budget or three consecutive non-promotions. Trigger-description evolution follows the same propose, evaluate, and retain-or-reject structure, but changes only $d_t$ and optimizes routing quality.

\begin{algorithm}[t]
\caption{Validation-Gated Harness Evolution}
\label{alg:evolution}
\begin{algorithmic}[1]
\STATE Evaluate initial Harness $H_0$ on $D_{\mathrm{val}}$
\STATE $H\leftarrow H_0$; $\mathcal{H}\leftarrow\emptyset$; $\mathcal{R}\leftarrow\emptyset$; $q\leftarrow0$
\FOR{$t=1,\ldots,R$}
  \STATE Collect current errors with failure types and protected correct cases
  \STATE Build feedback with these cases, prior-candidate errors $\mathcal{R}$, and history $\mathcal{H}$
  \STATE Propose restricted task-layer edits to obtain $H'$
  \STATE Evaluate $H'$ on the same validation data
  \IF{$\operatorname{Promote}(H',H)=1$}
    \STATE $H\leftarrow H'$; $q\leftarrow0$
  \ELSE
    \STATE Reject $H'$ and retain $H$; $q\leftarrow q+1$
  \ENDIF
  \STATE Append the diff, metric changes, and decision reason to $\mathcal{H}$
  \STATE $\mathcal{H}\leftarrow\operatorname{Tail}_3(\mathcal{H})$; $\mathcal{R}\leftarrow$ new errors introduced by $H'$
  \IF{$q=3$}
    \STATE \textbf{break}
  \ENDIF
\ENDFOR
\STATE Generate Maker--Checker review materials for the selected Harness
\STATE Freeze the selected Harness and evaluate $D_{\mathrm{test}}$
\end{algorithmic}
\end{algorithm}

All candidate selection uses development data. The held-out recognition test split and held-out trigger test split are evaluated only after candidate selection, and their errors are not returned to the loop. We repeat the procedure with seeds 42, 43, and 44, which control sampling, ordering, and baseline randomness; external API execution may still vary. The archived artifacts preserve accepted and rejected trajectories, allowing reported changes to be traced to concrete Harness versions.

\section{Experiments}

\subsection{Benchmark Construction and Quality Control}

The main benchmark contains 8,328 Chinese multi-turn dialogues from social-support and knowledge-seeking settings. The social-support portion contains 4,338 dialogues synthesized from publicly redistributed, anonymized PsyQA help-seeking seeds \cite{sun2021psyqa}. The knowledge-seeking portion contains 3,990 dialogues generated from the topic structures of 24 K--12 textbooks and 15 adult curricula or examinations. The evaluation unit is a complete dialogue. Life-stage evidence may be distributed across turns, delayed until later in the exchange, or mixed with cues pointing in the opposite direction.

Adult examples are deliberately difficult confounders. Social-domain negatives include young-adult help-seeking situations that resemble adolescent concerns in topic and style. Knowledge-domain negatives include university study, professional qualification, and adult examination settings that overlap lexically with school-related conversations. Systems must therefore bind evidence to the current speaker and distinguish educational content from the speaker's likely life stage.

Construction includes seed-level manual curation and generated-dialogue automated checks. Social-domain seeds undergo rule-based filtering followed by single-review manual curation before generation. Knowledge-domain labels are derived programmatically from curriculum stage and examination type. After generation, automated checks verify required fields, age--stage consistency, temporal formats, and stratum counts. This separation of generation, assessment, and downstream use follows established accounts of LLM-based data annotation and synthesis \cite{tan2024dataannotation}. We retain construction and split records following the broader rationale of transparent dataset documentation \cite{pushkarna2022datacards}.

The final split contains 5,830 training, 833 validation, and 1,665 test dialogues. The protocol introduces a deliberate evidence shift: all training and validation dialogues use explicit life-stage anchors, whereas all test dialogues require implicit inference without a direct age or stage statement. The test set contains 833 minor and 832 adult examples. Its scenario families and coarse families are disjoint from both training and validation, so the test set evaluates performance under a family-disjoint explicit-to-implicit shift within this controlled benchmark. The training split supports parameter estimation and forms the retrieval case base. Neither validation nor test examples enter retrieval. Validation data selects prompts, contracts, and Harness versions; the test split is evaluated after these artifacts are frozen.

Trigger-Eval is constructed separately because invocation and downstream classification answer different questions. It contains 480 requests balanced between positive and negative labels and covers direct requests and window-scanning conditions. We use a 288-example development split to revise and select the invocation description. The remaining 192 examples form a held-out trigger test split and are not used during description selection. The human study uses a 40-item challenge subset from the main test split: 10 Easy, 15 Hard, 10 Adversarial, and 5 Boundary items. Each of 50 non-expert participants labels all 40 dialogues and reports confidence on a five-point scale.

\subsection{Comparison Protocol}

All methods use the same minor/adult label space and partition. Trainable encoders and safeguards use the training set for parameter estimation and validation data for checkpoint selection. Prompt-based systems fix prompts, output contracts, and parsing rules on validation data. Minor-class average precision (AP) is computed from frozen per-example scores with tied scores evaluated jointly. Discriminative models use their positive-class posterior, while LLM systems use the structured \texttt{minor\_confidence} field; thresholded metrics use the returned Boolean label. We report minor-class precision, recall, and F1 over seeds 42, 43, and 44. The sample standard deviations summarize variation across these archived runs, including stochastic optimization and remote API execution, rather than sampling uncertainty over new dialogue populations.

The baselines cover three paradigms. Chinese-RoBERTa-wwm-ext represents supervised encoder fine-tuning \cite{cui2020chinese}. Aegis-Guard FT adapts the Aegis 1.0 defensive safeguard checkpoint \cite{ghosh2024aegis,inan2023llamaguard} to the binary task. Gemini Direct uses the same \texttt{gemini-2.5-flash-lite} classifier, dialogue window, JSON response mode, and confidence semantics as our method \cite{geminiteam2023gemini}. Its task-aware prompt covers speaker binding and minor/adult evidence, but it does not receive the Harness evidence package, retrieved cases, temporal features, schema-repair path, or version-evolution procedure. It is therefore a strong direct-prompt comparison for the additional system structure. We also evaluate the fine-tuned Aegis backbone with the full Harness. Because the Aegis and Gemini settings use different training and runtime protocols, the portability result is secondary to the shared-model comparison.

\begin{table*}[t]
\centering
\begin{tabular*}{\textwidth}{@{\extracolsep{\fill}}lcccc@{}}
\toprule
Method & AP $\uparrow$ & Minor F1 $\uparrow$ & Minor precision $\uparrow$ & Minor recall $\uparrow$ \\
\midrule
Chinese-RoBERTa-wwm-ext & $0.8154\pm0.0093$ & $0.6544\pm0.0179$ & $0.6743\pm0.0186$ & $0.6359\pm0.0222$ \\
Gemini Direct & $0.8815\pm0.0006$ & $0.8561\pm0.0010$ & $0.8317\pm0.0019$ & $0.8820\pm0.0007$ \\
Aegis-Guard FT & $0.8424\pm0.0045$ & $0.6640\pm0.0121$ & $0.8963\pm0.0200$ & $0.5278\pm0.0215$ \\
Aegis-Guard FT + Harness & $0.8744\pm0.0131$ & $0.7851\pm0.0178$ & $0.8541\pm0.0113$ & $0.7271\pm0.0331$ \\
Ours & $\mathbf{0.9574\pm0.0062}$ & $\mathbf{0.9198\pm0.0015}$ & $\mathbf{0.9360\pm0.0122}$ & $\mathbf{0.9044\pm0.0109}$ \\
\bottomrule
\end{tabular*}
\caption{Held-out main-task results on 1,665 dialogues over three runs. Values are mean $\pm$ sample standard deviation.}
\label{tab:main}
\end{table*}

\begin{table*}[t]
\centering
\small
\begin{tabular*}{\textwidth}{@{\extracolsep{\fill}}lccccc@{}}
\toprule
Variant & Acc. $\uparrow$ & AP $\uparrow$ & F1$_m$ $\uparrow$ & Prec.$_m$ $\uparrow$ & Rec.$_m$ $\uparrow$ \\
\midrule
Ours (selected Harness) & $\mathbf{0.9211\pm0.0019}$ & $\mathbf{0.9574\pm0.0062}$ & $\mathbf{0.9198\pm0.0015}$ & $0.9360\pm0.0122$ & $0.9044\pm0.0109$ \\
Ours (initial Harness) & $0.8975\pm0.0050$ & $0.9429\pm0.0036$ & $0.8971\pm0.0048$ & $0.9011\pm0.0073$ & $0.8932\pm0.0036$ \\
w/o contextual support & $0.8907\pm0.0033$ & $0.9246\pm0.0065$ & $0.8941\pm0.0037$ & $0.8672\pm0.0011$ & $\mathbf{0.9228\pm0.0077}$ \\
w/o evidence traceability & $0.9045\pm0.0026$ & $0.9434\pm0.0028$ & $0.9002\pm0.0029$ & $\mathbf{0.9434\pm0.0015}$ & $0.8607\pm0.0042$ \\
\bottomrule
\end{tabular*}
\caption{Held-out main-task ablations over three runs. Values are mean $\pm$ sample standard deviation.}
\label{tab:ablation}
\end{table*}

Ablations retain the base-model runtime, split, and scoring code. The initial Harness removes the effect of version evolution while preserving the initial structure. The context ablation removes train-derived retrieval and auxiliary temporal or scenario support. The traceability ablation removes the requirement that decisive cues map back to locatable dialogue evidence. Candidate selection uses validation data; all ablation numbers are computed on the held-out test split.

For the human reference, individual metrics use all 40 items. Strict majority resolves 38 items; two 25--25 ties remain unresolved. We report rater variation and agreement with Fleiss' $\kappa$ \cite{fleiss1971agreement}. Paired system comparisons and tie-sensitivity analyses are reported in the supplement because the challenge set contains only 40 independent items.

\section{Results}

\subsection{Main Task Performance}

Table~\ref{tab:main} reports the held-out main-task results. Our method obtains an AP of $0.9574\pm0.0062$ and a minor-class F1 of $0.9198\pm0.0015$, the highest values among the evaluated methods. Relative to Gemini Direct, AP increases by 7.60 percentage points and F1 by 6.37 points. Precision rises by 10.43 points, from 0.8317 to 0.9360, while recall rises by 2.24 points, from 0.8820 to 0.9044. The larger precision gain indicates that the framework mainly reduces adult false positives without sacrificing minor coverage. For each seed, paired item-bootstrap 95\% intervals exclude zero for both AP and F1 gains; exact McNemar tests on correctness give $p<6\times10^{-12}$ in all three runs (supplement).

The safeguard comparison shows a smaller but substantial gain. Aegis-Guard FT is precise but misses many positive cases, with recall 0.5278 and F1 0.6640. Adding the Harness raises recall to 0.7271 and F1 to 0.7851, while precision decreases from 0.8963 to 0.8541. The same pattern appears with a separately fine-tuned safeguard backbone, providing secondary evidence that the Harness is useful beyond the Gemini runtime. The different protocols preclude a broad cross-family ranking.

\subsection{Component Ablations}

Table~\ref{tab:ablation} shows different precision--recall effects. Removing contextual support raises recall from 0.9044 to 0.9228 but lowers precision from 0.9360 to 0.8672, with losses of 3.28 AP points and 2.57 F1 points. Context therefore contributes mainly by suppressing false minor predictions on adult examples that resemble youth dialogue. This ablation removes retrieval together with temporal and scenario support, so it does not isolate retrieval alone. Removing evidence traceability raises precision slightly to 0.9434 but lowers recall to 0.8607; F1 and AP fall by 1.96 and 1.40 points. Requiring locatable, typed evidence helps retain positive cues that are weak in isolation or distributed across turns.

The selected Harness also improves over its structurally matched initial version. F1 rises from 0.8971 to 0.9198 and AP from 0.9429 to 0.9574, corresponding to gains of 2.27 and 1.45 points. Precision increases from 0.9011 to 0.9360 and recall from 0.8932 to 0.9044. These joint gains are consistent with revisions informed by current errors, protected correct cases, and recent failed updates. The paired F1 interval excludes zero in every run; the AP interval does so in two of three runs. Version selection occurs on validation data before final test evaluation.

\subsection{Trigger Boundary and Version Governance}

\begin{table}[t]
\centering
\small
\setlength{\tabcolsep}{1mm}
\begin{tabular}{lccc}
\toprule
Description & Prec. & Rec. & F1 \\
\midrule
Initial & $0.831\pm0.012$ & $0.715\pm0.042$ & $0.768\pm0.019$ \\
Selected & $0.823\pm0.031$ & $\mathbf{0.861\pm0.043}$ & $\mathbf{0.841\pm0.006}$ \\
\bottomrule
\end{tabular}
\caption{Initial and development-selected trigger descriptions on the held-out trigger test split over three runs (mean $\pm$ sample standard deviation).}
\label{tab:trigger}
\end{table}

Table~\ref{tab:trigger} shows that trigger evolution primarily expands coverage. Recall rises by 14.58 points, while precision decreases by 0.82 points, producing a 7.23-point F1 gain. Invocation success rises from $0.9983\pm0.0030$ to $1.0000\pm0.0000$. This result supports separate optimization of routing behavior and downstream recognition.

\begin{figure}[t]
\centering
\includegraphics[width=\columnwidth]{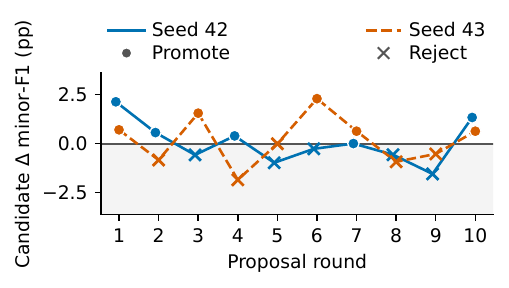}
\caption{Candidate minor-F1 change relative to the current incumbent for the two full-budget validation runs. Filled circles denote promoted candidates and crosses denote rejected candidates.}
\label{fig:evolution}
\end{figure}

Figure~\ref{fig:evolution} shows that candidate quality is non-monotonic: several proposals fall below the current incumbent and are rejected, while accepted proposals update the comparison point for later rounds. The two plotted runs reached the full ten-round budget; complete records for all runs are provided in the supplement. Across those records, 11 of 25 candidates are promoted and 14 are rejected. A rejection is retained as negative experience: its diff, metric changes, decision reason, and new errors are available to subsequent proposals. The trajectories report validation-time proposal behavior. Held-out results in Table~\ref{tab:ablation} provide a separate check: the selected Harness improves over its initial version by 2.27 F1 points and 1.45 AP points.

\subsection{Human Challenge-Set Reference}

On the difficulty-enriched 40-item subset, 50 non-expert adult readers achieve $72.6\%$ mean accuracy and $72.8\%$ mean minor-F1, with Fleiss' $\kappa=0.405$. Accuracy falls to $54.8\%$ on Boundary and $65.7\%$ on Hard items, compared with $80.8\%$ on Adversarial and $83.4\%$ on Easy items. Together with the trigger results, this pattern separates two sources of difficulty: deciding whether age inference is warranted and resolving weak or disputed evidence after invocation. Participant variation, confidence, majority-vote comparisons, paired system comparisons, and tie sensitivity are reported in the supplement.

\section{Discussion}

The results support a task design that separates invocation from recognition while retaining speaker-bound evidence across the full dialogue. Contextual support and evidence traceability affect different parts of the precision--recall trade-off. The held-out initial-to-selected comparison also shows that validation-guided Harness revision can improve the initial version under the stated admission rules. Trigger routing remains a separate source of error: a recognition model cannot help when age inference is invoked for the wrong request or not invoked when it is needed.

\subsection{Limitations}

The current evidence focuses on implicit minor detection and a controlled dialogue-level synthetic benchmark. It therefore supports the effectiveness of the proposed method under the reported conditions, without directly establishing performance in real platform deployments. Structured constraints, retrieval support, and Maker--Checker review introduce additional system cost while providing auditability, boundary stability, and version governance. Evaluation under more realistic deployment conditions, cross-lingual validation, and performance retention in lower-cost configurations remain directions for future work.

\section{Conclusion}

We introduced implicit minor detection as a speaker-bound multi-turn task and constructed an 8,328-dialogue benchmark with a separate trigger-boundary evaluation. Our framework combines a fixed LLM, a versioned Harness, and Maker--Checker review. Harness evolution uses current errors together with protected correct cases and the recorded outcomes of recent accepted and rejected updates. On the held-out main task, the framework improves AP by 7.60 percentage points and minor-class F1 by 6.37 points over direct inference with the same LLM family. Trigger evolution adds 7.23 F1 points, showing that invocation boundaries can be improved separately from recognition.

\section{Ethical Statement}

This work treats minor status as a risk assessment grounded in multi-turn dialogue evidence. The assessment is designed to support protective decisions and does not verify a user's legal age. Inferring a sensitive attribute from indirect language creates privacy and misuse risks \cite{staab2024privacy,weidinger2022risks}. The output should therefore not be used alone to trigger high-stakes automated actions. High-uncertainty and boundary cases should retain access to human review and a correction or appeal process. We position the framework as an auxiliary risk-control layer that can operate alongside platform-level age assurance, human judgment, and formal identity-verification mechanisms.

The human challenge-set study used a voluntary anonymous questionnaire completed by 50 adults aged 20--30. Participants were informed that some items contain psychological distress or self-harm expressions and consented to research use upon submission. No direct identifiers were collected, and the paper reports only aggregate and item-level statistics.

\bibliography{paper}

\begin{thebibliography}{35}
\providecommand{\natexlab}[1]{#1}

\bibitem[{Bai et~al.(2024)Bai, Liu, Bu, He, Liu, Zhou, Lin, Su, Ge, Zheng, and
  Ouyang}]{bai2024mtbench101}
Bai, G.; Liu, J.; Bu, X.; He, Y.; Liu, J.; Zhou, Z.; Lin, Z.; Su, W.; Ge, T.;
  Zheng, B.; and Ouyang, W. 2024.
\newblock {MT-Bench-101}: A Fine-Grained Benchmark for Evaluating Large
  Language Models in Multi-Turn Dialogues.
\newblock In \emph{Proceedings of the 62nd Annual Meeting of the Association
  for Computational Linguistics}, 7421--7454. Association for Computational
  Linguistics.

\bibitem[{Cui et~al.(2020)Cui, Che, Liu, Qin, Wang, and Hu}]{cui2020chinese}
Cui, Y.; Che, W.; Liu, T.; Qin, B.; Wang, S.; and Hu, G. 2020.
\newblock Revisiting Pre-Trained Models for Chinese Natural Language
  Processing.
\newblock In \emph{Findings of the Association for Computational Linguistics:
  EMNLP 2020}, 657--668. Association for Computational Linguistics.

\bibitem[{DeYoung et~al.(2020)DeYoung, Jain, Rajani, Lehman, Xiong, Socher, and
  Wallace}]{deyoung2020eraser}
DeYoung, J.; Jain, S.; Rajani, N.~F.; Lehman, E.; Xiong, C.; Socher, R.; and
  Wallace, B.~C. 2020.
\newblock {ERASER}: A Benchmark to Evaluate Rationalized {NLP} Models.
\newblock In \emph{Proceedings of the 58th Annual Meeting of the Association
  for Computational Linguistics}, 4443--4458. Association for Computational
  Linguistics.

\bibitem[{Dhuliawala et~al.(2024)Dhuliawala, Komeili, Xu, Raileanu, Li,
  {\c{C}}elikyilmaz, and Weston}]{dhuliawala2024chainverification}
Dhuliawala, S.; Komeili, M.; Xu, J.; Raileanu, R.; Li, X.; {\c{C}}elikyilmaz,
  A.; and Weston, J. 2024.
\newblock Chain-of-Verification Reduces Hallucination in Large Language Models.
\newblock In \emph{Findings of the Association for Computational Linguistics:
  ACL 2024}, 3563--3578. Association for Computational Linguistics.

\bibitem[{Fernando et~al.(2024)Fernando, Banarse, Michalewski, Osindero, and
  Rockt{\"a}schel}]{fernando2024promptbreeder}
Fernando, C.; Banarse, D.~S.; Michalewski, H.; Osindero, S.; and
  Rockt{\"a}schel, T. 2024.
\newblock Promptbreeder: Self-Referential Self-Improvement via Prompt
  Evolution.
\newblock In \emph{Proceedings of the 41st International Conference on Machine
  Learning}, volume 235 of \emph{Proceedings of Machine Learning Research},
  13481--13544. PMLR.

\bibitem[{Fleiss(1971)}]{fleiss1971agreement}
Fleiss, J.~L. 1971.
\newblock Measuring Nominal Scale Agreement among Many Raters.
\newblock \emph{Psychological Bulletin}, 76(5): 378--382.

\bibitem[{{Gemini Team}(2023)}]{geminiteam2023gemini}
{Gemini Team}. 2023.
\newblock Gemini: A Family of Highly Capable Multimodal Models.
\newblock arXiv:2312.11805.

\bibitem[{Ghosh et~al.(2024)Ghosh, Varshney, Galinkin, and
  Parisien}]{ghosh2024aegis}
Ghosh, S.; Varshney, P.; Galinkin, E.; and Parisien, C. 2024.
\newblock {AEGIS}: Online Adaptive {AI} Content Safety Moderation with Ensemble
  of {LLM} Experts.
\newblock \emph{arXiv preprint arXiv:2404.05993}.

\bibitem[{Gou et~al.(2024)Gou, Shao, Gong, Shen, Yang, Duan, and
  Chen}]{gou2024critic}
Gou, Z.; Shao, Z.; Gong, Y.; Shen, Y.; Yang, Y.; Duan, N.; and Chen, W. 2024.
\newblock {CRITIC}: Large Language Models Can Self-Correct with
  Tool-Interactive Critiquing.
\newblock In \emph{The Twelfth International Conference on Learning
  Representations}.

\bibitem[{Han et~al.(2024)Han, Rao, Ettinger, Jiang, Lin, Lambert, Choi, and
  Dziri}]{han2024wildguard}
Han, S.; Rao, K.; Ettinger, A.; Jiang, L.; Lin, B.~Y.; Lambert, N.; Choi, Y.;
  and Dziri, N. 2024.
\newblock WildGuard: Open One-stop Moderation Tools for Safety Risks,
  Jailbreaks, and Refusals of {LLM}s.
\newblock In \emph{Advances in Neural Information Processing Systems 37},
  8093--8131. Neural Information Processing Systems Foundation, Inc.

\bibitem[{Inan et~al.(2023)Inan, Upasani, Chi, Rungta, Iyer, Mao, Tontchev, Hu,
  Fuller, Testuggine, and Khabsa}]{inan2023llamaguard}
Inan, H.; Upasani, K.; Chi, J.; Rungta, R.; Iyer, K.; Mao, Y.; Tontchev, M.;
  Hu, Q.; Fuller, B.; Testuggine, D.; and Khabsa, M. 2023.
\newblock Llama Guard: {LLM}-Based Input-Output Safeguard for Human-{AI}
  Conversations.
\newblock arXiv:2312.06674.

\bibitem[{Jarvie and Renaud(2024)}]{jarvie2024ageverification}
Jarvie, C.; and Renaud, K. 2024.
\newblock Online Age Verification: Government Legislation, Supplier
  Responsibilization, and Public Perceptions.
\newblock \emph{Children}, 11(9): 1068.

\bibitem[{Khattab et~al.(2024)Khattab, Singhvi, Maheshwari, Zhang, Santhanam,
  Vardhamanan, Haq, Sharma, Joshi, Moazam, Miller, Zaharia, and
  Potts}]{khattab2023dspy}
Khattab, O.; Singhvi, A.; Maheshwari, P.; Zhang, Z.; Santhanam, K.;
  Vardhamanan, S.; Haq, S.; Sharma, A.; Joshi, T.~T.; Moazam, H.; Miller, H.;
  Zaharia, M.; and Potts, C. 2024.
\newblock {DSPy}: Compiling Declarative Language Model Calls into
  State-of-the-Art Pipelines.
\newblock In \emph{The Twelfth International Conference on Learning
  Representations}.

\bibitem[{Lewis et~al.(2020)Lewis, Perez, Piktus, Petroni, Karpukhin, Goyal,
  Kuttler, Lewis, Yih, Rockt{\"a}schel, Riedel, and Kiela}]{lewis2020rag}
Lewis, P.; Perez, E.; Piktus, A.; Petroni, F.; Karpukhin, V.; Goyal, N.;
  Kuttler, H.; Lewis, M.; Yih, W.-t.; Rockt{\"a}schel, T.; Riedel, S.; and
  Kiela, D. 2020.
\newblock Retrieval-Augmented Generation for Knowledge-Intensive {NLP} Tasks.
\newblock In \emph{Advances in Neural Information Processing Systems 33},
  9459--9474.

\bibitem[{Li, Song, and Fan(2022)}]{li2022privatepersona}
Li, H.; Song, Y.; and Fan, L. 2022.
\newblock You Don't Know My Favorite Color: Preventing Dialogue Representations
  from Revealing Speakers' Private Personas.
\newblock In \emph{Proceedings of the 2022 Conference of the North American
  Chapter of the Association for Computational Linguistics: Human Language
  Technologies}, 5858--5870. Association for Computational Linguistics.

\bibitem[{Livingstone et~al.(2024)Livingstone, Nair, Stoilova, van~der Hof, and
  {\c{C}}a{\u{g}}lar}]{livingstone2024ageassurance}
Livingstone, S.; Nair, A.; Stoilova, M.; van~der Hof, S.; and
  {\c{C}}a{\u{g}}lar, C. 2024.
\newblock Children's Rights and Online Age Assurance Systems.
\newblock \emph{The International Journal of Children's Rights}, 32(3):
  721--747.

\bibitem[{Madaan et~al.(2023)Madaan, Tandon, Gupta, Hallinan, Gao, Wiegreffe,
  Alon, Dziri, Prabhumoye, Yang, Gupta, Majumder, Hermann, Welleck,
  Yazdanbakhsh, and Clark}]{madaan2023selfrefine}
Madaan, A.; Tandon, N.; Gupta, P.; Hallinan, S.; Gao, L.; Wiegreffe, S.; Alon,
  U.; Dziri, N.; Prabhumoye, S.; Yang, Y.; Gupta, S.; Majumder, B.~P.; Hermann,
  K.; Welleck, S.; Yazdanbakhsh, A.; and Clark, P. 2023.
\newblock Self-Refine: Iterative Refinement with Self-Feedback.
\newblock In \emph{Advances in Neural Information Processing Systems 36},
  46534--46594. Neural Information Processing Systems Foundation, Inc.

\bibitem[{Pushkarna, Zaldivar, and Kjartansson(2022)}]{pushkarna2022datacards}
Pushkarna, M.; Zaldivar, A.; and Kjartansson, O. 2022.
\newblock Data Cards: Purposeful and Transparent Dataset Documentation for
  Responsible {AI}.
\newblock In \emph{Proceedings of the 2022 ACM Conference on Fairness,
  Accountability, and Transparency}, 1776--1826. ACM.

\bibitem[{Ribeiro et~al.(2020)Ribeiro, Wu, Guestrin, and
  Singh}]{ribeiro2020checklist}
Ribeiro, M.~T.; Wu, T.; Guestrin, C.; and Singh, S. 2020.
\newblock Beyond Accuracy: Behavioral Testing of {NLP} Models with {CheckList}.
\newblock In \emph{Proceedings of the 58th Annual Meeting of the Association
  for Computational Linguistics}, 4902--4912. Association for Computational
  Linguistics.

\bibitem[{Rosso(2017)}]{rosso2017authorprofiling}
Rosso, P. 2017.
\newblock Author Profiling at {PAN}: From Age and Gender Identification to
  Language Variety Identification (Invited Talk).
\newblock In \emph{Proceedings of the Fourth Workshop on NLP for Similar
  Languages, Varieties and Dialects (VarDial)}, 46--46. Association for
  Computational Linguistics.

\bibitem[{R\"ottger et~al.(2024)R\"ottger, Kirk, Vidgen, Attanasio, Bianchi,
  and Hovy}]{rottger2024xstest}
R\"ottger, P.; Kirk, H.; Vidgen, B.; Attanasio, G.; Bianchi, F.; and Hovy, D.
  2024.
\newblock {XSTest}: A Test Suite for Identifying Exaggerated Safety Behaviours
  in Large Language Models.
\newblock In \emph{Proceedings of the 2024 Conference of the North American
  Chapter of the Association for Computational Linguistics: Human Language
  Technologies}, 5377--5400. Association for Computational Linguistics.

\bibitem[{Shinn et~al.(2023)Shinn, Cassano, Gopinath, Narasimhan, and
  Yao}]{shinn2023reflexion}
Shinn, N.; Cassano, F.; Gopinath, A.; Narasimhan, K.; and Yao, S. 2023.
\newblock Reflexion: Language Agents with Verbal Reinforcement Learning.
\newblock In \emph{Advances in Neural Information Processing Systems 36},
  8634--8652. Neural Information Processing Systems Foundation, Inc.

\bibitem[{Staab et~al.(2024)Staab, Vero, Balunovi{\'c}, and
  Vechev}]{staab2024privacy}
Staab, R.; Vero, M.; Balunovi{\'c}, M.; and Vechev, M. 2024.
\newblock Beyond Memorization: Violating Privacy via Inference with Large
  Language Models.
\newblock In \emph{The Twelfth International Conference on Learning
  Representations}.

\bibitem[{Sun et~al.(2021)Sun, Lin, Zheng, Liu, and Huang}]{sun2021psyqa}
Sun, H.; Lin, Z.; Zheng, C.; Liu, S.; and Huang, M. 2021.
\newblock {P}sy{QA}: A {C}hinese Dataset for Generating Long Counseling Text
  for Mental Health Support.
\newblock In \emph{Findings of the Association for Computational Linguistics:
  ACL-IJCNLP 2021}, 1489--1503. Online: Association for Computational
  Linguistics.

\bibitem[{Tan et~al.(2024)Tan, Li, Wang, Beigi, Jiang, Bhattacharjee, Karami,
  Li, Cheng, and Liu}]{tan2024dataannotation}
Tan, Z.; Li, D.; Wang, S.; Beigi, A.; Jiang, B.; Bhattacharjee, A.; Karami, M.;
  Li, J.; Cheng, L.; and Liu, H. 2024.
\newblock Large Language Models for Data Annotation and Synthesis: A Survey.
\newblock In \emph{Proceedings of the 2024 Conference on Empirical Methods in
  Natural Language Processing}, 930--957. Association for Computational
  Linguistics.

\bibitem[{Tigunova et~al.(2020)Tigunova, Yates, Mirza, and
  Weikum}]{tigunova2020charm}
Tigunova, A.; Yates, A.; Mirza, P.; and Weikum, G. 2020.
\newblock {CHARM}: Inferring Personal Attributes from Conversations.
\newblock In \emph{Proceedings of the 2020 Conference on Empirical Methods in
  Natural Language Processing (EMNLP)}, 5391--5404. Association for
  Computational Linguistics.

\bibitem[{Wang et~al.(2022)Wang, Zhou, Koncel-Kedziorski, Marin, and
  Xia}]{wang2022personalattributes}
Wang, Z.; Zhou, X.; Koncel-Kedziorski, R.; Marin, A.; and Xia, F. 2022.
\newblock Extracting and Inferring Personal Attributes from Dialogue.
\newblock In \emph{Proceedings of the 4th Workshop on NLP for Conversational
  AI}, 58--69. Association for Computational Linguistics.

\bibitem[{Weidinger et~al.(2022)Weidinger, Uesato, Rauh, Griffin, Huang,
  Mellor, Glaese, Cheng, Balle, Kasirzadeh, Biles, Brown, Kenton, Hawkins,
  Stepleton, Birhane, Hendricks, Rimell, Isaac, Haas, Legassick, Irving, and
  Gabriel}]{weidinger2022risks}
Weidinger, L.; Uesato, J.; Rauh, M.; Griffin, C.; Huang, P.-S.; Mellor, J.;
  Glaese, A.; Cheng, M.; Balle, B.; Kasirzadeh, A.; Biles, C.; Brown, S.;
  Kenton, Z.; Hawkins, W.; Stepleton, T.; Birhane, A.; Hendricks, L.~A.;
  Rimell, L.; Isaac, W.; Haas, J.; Legassick, S.; Irving, G.; and Gabriel, I.
  2022.
\newblock Taxonomy of Risks Posed by Language Models.
\newblock In \emph{Proceedings of the 2022 ACM Conference on Fairness,
  Accountability, and Transparency}, 214--229. ACM.

\bibitem[{Yang et~al.(2024)Yang, Wang, Lu, Liu, Le, Zhou, and
  Chen}]{yang2024opro}
Yang, C.; Wang, X.; Lu, Y.; Liu, H.; Le, Q.~V.; Zhou, D.; and Chen, X. 2024.
\newblock Large Language Models as Optimizers.
\newblock In \emph{The Twelfth International Conference on Learning
  Representations}.

\bibitem[{Yao et~al.(2023)Yao, Zhao, Yu, Du, Shafran, Narasimhan, and
  Cao}]{yao2023react}
Yao, S.; Zhao, J.; Yu, D.; Du, N.; Shafran, I.; Narasimhan, K.; and Cao, Y.
  2023.
\newblock {ReAct}: Synergizing Reasoning and Acting in Language Models.
\newblock In \emph{The Eleventh International Conference on Learning
  Representations}.

\bibitem[{Yu(2025)}]{yu2025safeguarding}
Yu, Y. 2025.
\newblock Safeguarding Children in Generative {AI}: Risk Frameworks and
  Parental Control Tools.
\newblock In \emph{The 2025 ACM International Conference on Supporting Group
  Work ({GROUP} Companion 2025)}, 121--123. ACM.

\bibitem[{Yu et~al.(2025{\natexlab{a}})Yu, Liu, Zhang, Huang, and
  Wang}]{yu2025yair}
Yu, Y.; Liu, Y.; Zhang, J.; Huang, Y.; and Wang, Y. 2025{\natexlab{a}}.
\newblock Youth-Centered {GAI} Risks ({YAIR}): A Taxonomy of Generative {AI}
  Risks from Empirical Data.
\newblock In \emph{Twenty-First Symposium on Usable Privacy and Security},
  149--165. USENIX Association.

\bibitem[{Yu et~al.(2025{\natexlab{b}})Yu, Liu, Zhang, Huang, and
  Wang}]{yu2025youthsafe}
Yu, Y.; Liu, Y.; Zhang, Y.; Huang, Y.; and Wang, Y. 2025{\natexlab{b}}.
\newblock YouthSafe: A Youth-Centric Safety Benchmark and Safeguard Model for
  Large Language Models.
\newblock In \emph{Proceedings of the 2025 ACM SIGSAC Conference on Computer
  and Communications Security}, 4349--4363. ACM.

\bibitem[{Y{\"u}ksekg{\"o}n{\"u}l et~al.(2024)Y{\"u}ksekg{\"o}n{\"u}l, Bianchi,
  Boen, Liu, Huang, Guestrin, and Zou}]{yuksekgonul2024textgrad}
Y{\"u}ksekg{\"o}n{\"u}l, M.; Bianchi, F.; Boen, J.; Liu, S.; Huang, Z.;
  Guestrin, C.; and Zou, J. 2024.
\newblock {TextGrad}: Automatic ``Differentiation'' via Text.
\newblock arXiv:2406.07496.

\bibitem[{Zhu et~al.(2023)Zhu, Li, Mao, Pandelea, and Cambria}]{zhu2023paed}
Zhu, L.; Li, W.; Mao, R.; Pandelea, V.; and Cambria, E. 2023.
\newblock {PAED}: Zero-Shot Persona Attribute Extraction in Dialogues.
\newblock In \emph{Proceedings of the 61st Annual Meeting of the Association
  for Computational Linguistics (Volume 1: Long Papers)}, 9771--9787.
  Association for Computational Linguistics.

\end{thebibliography}
\clearpage
\appendix
\setcounter{secnumdepth}{2}
\setcounter{section}{0}
\setcounter{table}{0}
\renewcommand{\thetable}{S\arabic{table}}

\section*{Technical Supplement for Implicit Minor Detection}
\subsection*{Supplement Overview}
This supplement expands the evidence reported in the main paper without changing its claims. Section~\ref{app:benchmark} documents the three evaluation surfaces and their isolation rules. Section~\ref{app:protocol} records the frozen runtime, optimization, and statistical protocol. Sections~\ref{app:evolution} and~\ref{app:trigger} report version evolution for recognition and trigger routing. Section~\ref{app:human} gives the complete human challenge-set protocol, agreement analysis, paired comparison, and tie sensitivity. The main paper remains self-contained; the material below provides audit detail and complete trajectories.

\section{Benchmark and Evaluation Protocol}
\label{app:benchmark}
\subsection{Main Recognition Benchmark}
The main benchmark contains 8,328 Chinese multi-turn dialogues: 4,338 social-support dialogues and 3,990 knowledge-seeking dialogues. The frozen split contains 5,830 training, 833 validation, and 1,665 test examples. The test set contains 833 minor and 832 adult examples. The retrieval case base is identical to the training split; validation and test examples are excluded from retrieval.

Social-support seeds pass rule-based filtering and single-review manual curation before dialogue generation. Knowledge-domain labels follow curriculum stage and examination type. Adult controls cover university study, professional examinations, employment, renting, and other responsibility structures that overlap lexically with school-related dialogue. These controls require the classifier to resolve the speaker and life stage instead of relying on isolated school or family terms.

Construction records include generation templates, automated field and label checks, stratum counts, split manifests, and duplicate audits. Generated-dialogue checks cover required fields, label and age--stage consistency, temporal formats, duplicate detection, stratum counts, and split isolation. Manual review applies to the pre-generation social seeds.

\begin{table*}[t]
\centering
\footnotesize
\begin{tabular*}{\textwidth}{@{\extracolsep{\fill}}lrrrl@{}}
\toprule
Evaluation surface & Total & Development & Held-out & Primary role \\
\midrule
Main recognition benchmark & 8,328 & 5,830 train + 833 validation & 1,665 test & Speaker-level minor recognition \\
Trigger-Eval & 480 & 288 development & 192 test & Invocation-boundary evaluation \\
Human challenge set & 40 & --- & 40 shared items & Difficulty and disagreement reference \\
\bottomrule
\end{tabular*}
\caption{Evaluation surfaces used in the paper. Trigger-Eval and the human challenge set are separate from recognition training and retrieval.}
\label{tab:s-surfaces}
\end{table*}

\subsection{Family-Disjoint Split Audit}
The split audit uses \emph{family} and \emph{coarse-family} keys. For social examples, family pairs the source label with the originating seed identifier; the coarse key is identical. For positive knowledge examples, family uses subject and chapter when chapter metadata are available, while the coarse key uses subject only. For negative knowledge examples, family uses the source file when available and otherwise the subject; the coarse key applies the same source-file-or-subject grouping. The frozen manifest reports zero family and coarse-family overlap between test and either training or validation.

\subsection{Trigger and Human Evaluation Sets}
Trigger-Eval contains 480 balanced requests covering direct requests and window-scanning conditions. Description selection uses 288 development examples; the remaining 192 requests form the held-out trigger test split. Trigger development changes only the routing description and does not update the recognition pipeline.

The human study uses a difficulty-enriched 40-item subset of the recognition test split: 10 Easy, 15 Hard, 10 Adversarial, and 5 Boundary items. The subset contains 22 minor and 18 adult labels. It is used only for the human reference and paired challenge-set analysis.

\section{Reproducibility and Statistical Protocol}
\label{app:protocol}
\subsection{Recognition Systems and Baselines}
The recognition classifier and Gemini Direct both use the \texttt{gemini-2.5-flash-lite} alias through the same OpenAI-compatible endpoint with JSON-object response mode. The endpoint did not return an immutable provider-side checkpoint identifier. Harness inference retrieves the top three cases with \texttt{text-embedding-3-small}. Gemini Direct uses a task-aware prompt covering speaker binding, the under-18 decision boundary, conflicting evidence, and confidence semantics. Its archived prompt hash is \texttt{5e0494c2...3971ff}; the full SHA-256 value is stored in the artifact manifest. Gemini Direct omits the Harness evidence package, retrieval, temporal features, schema repair, and development loop.

Chinese-RoBERTa uses \path{hfl/chinese-roberta-wwm-ext}, a maximum sequence length of 512, three epochs, learning rate $2\times10^{-5}$, train/evaluation batch sizes 16/32, weight decay 0.01, and warmup ratio 0.06. Aegis-Guard FT uses \path{nvidia/Aegis-AI-Content-Safety-LlamaGuard-Defensive-1.0}, a maximum sequence length of 1024, one epoch, learning rate $2\times10^{-5}$, batch size 1 with gradient accumulation 8, and warmup ratio 0.03. The archived records do not contain an immutable revision for the locally cached Aegis backbone.

\subsection{Harness Evolution and Version Review}
The Harness optimizer uses \texttt{DeepSeek-V3}, a maximum search budget of ten rounds, and patience three. At each round, it receives the incumbent's errors and failure types, correct validation cases to protect, new errors introduced by the preceding candidate, and the diffs, metric changes, and acceptance or rejection reasons from the three most recent rounds. The base-model weights, output schema, controller, and evaluation code remain fixed.

After validation selection, the implementation produces a version-diff report and an eight-item Maker--Checker checklist for each run. The checklist covers changed predictions, low-confidence and boundary-sensitive cases, label credibility, practical acceptability, boundary drift, explanation risk, the evaluation record, and edited prompt text. The selected versions and review materials underwent manual inspection before freezing. This review is a version-governance procedure, not an additional performance measurement.

Main results use seeds 42, 43, and 44. The seeds control sampling, ordering, and trainable-baseline initialization. External API execution is not bitwise deterministic, so the reported sample standard deviation describes the three archived executions rather than uncertainty over a new dialogue population. Frozen artifacts retain schemas, prompt resources, parsing rules, version decisions, prediction files, and metric reports.

\subsection{Metrics and Paired Inference}
Minor-class AP is recomputed from frozen per-example scores with scikit-learn's average-precision implementation. Tied scores are evaluated jointly, making the result invariant to input order. F1, precision, recall, and accuracy use the returned Boolean label.

Table~\ref{tab:s-paired} reports percentile intervals from 2,000 paired, gold-stratified item-bootstrap replicates. Each replicate applies the same sampled indices to both systems. McNemar tests use paired item-level correctness. Intervals and tests are computed separately for each archived seed and are reported without multiplicity adjustment.

\begin{table}[H]
\centering
\scriptsize
\setlength{\tabcolsep}{2pt}
\begin{tabular}{lccc}
\toprule
Seed & $\Delta$AP [95\% CI] & $\Delta$F1$_m$ [95\% CI] & McNemar $p$ \\
\midrule
\multicolumn{4}{l}{Selected vs. Gemini Direct} \\
42 & $.0795\;[.0601,.0999]$ & $.0644\;[.0460,.0841]$ & $3.16\times10^{-13}$ \\
43 & $.0686\;[.0491,.0891]$ & $.0613\;[.0435,.0784]$ & $2.97\times10^{-12}$ \\
44 & $.0798\;[.0614,.0986]$ & $.0655\;[.0475,.0823]$ & $5.84\times10^{-14}$ \\
\midrule
\multicolumn{4}{l}{Selected vs. Initial Harness} \\
42 & $.0212\;[.0067,.0347]$ & $.0279\;[.0115,.0436]$ & $.000327$ \\
43 & $.0033\;[-.0102,.0173]$ & $.0177\;[.0024,.0324]$ & $.0320$ \\
44 & $.0190\;[.0059,.0318]$ & $.0226\;[.0070,.0379]$ & $.00316$ \\
\bottomrule
\end{tabular}
\caption{Paired uncertainty for the main shared-model comparison and the initial-to-selected Harness comparison. Intervals are computed separately for each archived run on the shared 1,665-item test set.}
\label{tab:s-paired}
\end{table}

For Selected versus Gemini Direct, every AP and F1 interval excludes zero, and every correctness comparison is significant under the exact McNemar test. For Selected versus Initial, F1 intervals exclude zero in all three runs. The AP interval crosses zero for seed 43 but excludes zero for seeds 42 and 44. This pattern supports the shared-model comparison most strongly and treats initial-to-selected improvement as a repeated held-out check rather than a uniform per-seed AP gain.

\section{Main-Task Harness Evolution}
\label{app:evolution}
The main paper plots candidate F1 change relative to the incumbent for seeds 42 and 43. Table~\ref{tab:s-evolution} first summarizes all three runs. Tables~\ref{tab:s-seed42}--\ref{tab:s-seed44} then report every candidate, including rejected proposals. F1, recall, and precision are validation metrics from the archived trajectory reports. Per-round confidence scores were not stored, so AP cannot be reconstructed for these trajectories.

\begin{table}[H]
\centering
\small
\begin{tabular}{lrrrc}
\toprule
Seed & Prop. & Prom. & Rej. & F1: initial $\to$ selected \\
\midrule
42 & 10 & 5 & 5 & $.9009\to.9455$ \\
43 & 10 & 5 & 5 & $.9067\to.9652$ \\
44 & 5 & 1 & 4 & $.9168\to.9483$ \\
\midrule
All & 25 & 11 & 14 & --- \\
\bottomrule
\end{tabular}
\caption{Summary of main-task validation evolution. Rejected candidates do not replace the incumbent.}
\label{tab:s-evolution}
\end{table}

Across the three trajectories, 11 of 25 proposals are promoted and 14 are rejected. Candidate quality is non-monotonic in every run. The history supplied to later proposals retains each candidate diff, metric change, decision reason, and newly introduced errors, so rejection records become explicit negative feedback rather than discarded attempts. Seed 44 reaches the patience criterion after three consecutive non-promotions; its selected incumbent remains the promoted v2 candidate.

\begin{table}[H]
\centering
\small
\begin{tabular}{llrrr}
\toprule
Version & Decision & F1 & Recall & Precision \\
\midrule
v0 & initial incumbent & .9009 & .8197 & 1.0000 \\
v1 & promote & .9223 & .8558 & 1.0000 \\
v2 & promote & .9280 & .8678 & .9972 \\
v3 & reject & .9223 & .8558 & 1.0000 \\
v4 & promote & .9320 & .8726 & 1.0000 \\
v5 & reject & .9223 & .8558 & 1.0000 \\
v6 & reject & .9294 & .8702 & .9972 \\
v7 & promote & .9321 & .8750 & .9973 \\
v8 & reject & .9265 & .8630 & 1.0000 \\
v9 & reject & .9167 & .8462 & 1.0000 \\
v10 & promote & .9455 & .8966 & 1.0000 \\
\bottomrule
\end{tabular}
\caption{Main-task validation trajectory for seed 42. A rejected candidate leaves the incumbent unchanged.}
\label{tab:s-seed42}
\end{table}

\begin{table}[H]
\centering
\small
\begin{tabular}{llrrr}
\toprule
Version & Decision & F1 & Recall & Precision \\
\midrule
v0 & initial incumbent & .9067 & .8293 & 1.0000 \\
v1 & promote & .9138 & .8413 & 1.0000 \\
v2 & reject & .9055 & .8293 & .9971 \\
v3 & promote & .9294 & .8702 & .9972 \\
v4 & reject & .9110 & .8365 & 1.0000 \\
v5 & reject & .9294 & .8702 & .9972 \\
v6 & promote & .9524 & .9135 & .9948 \\
v7 & promote & .9588 & .9231 & .9974 \\
v8 & reject & .9496 & .9062 & .9974 \\
v9 & reject & .9536 & .9135 & .9974 \\
v10 & promote & .9652 & .9327 & 1.0000 \\
\bottomrule
\end{tabular}
\caption{Main-task validation trajectory for seed 43. A rejected candidate leaves the incumbent unchanged.}
\label{tab:s-seed43}
\end{table}

\begin{table}[H]
\centering
\small
\begin{tabular}{llr}
\toprule
Version & Decision & F1 \\
\midrule
v0 & initial incumbent & .9168 \\
v1 & reject & .9156 \\
v2 & promote & .9483 \\
v3 & reject & .9430 \\
v4 & reject & .9388 \\
v5 & reject & .9430 \\
\bottomrule
\end{tabular}
\caption{Main-task validation trajectory for seed 44. The run stops after rounds 3--5 yield three consecutive non-promotions.}
\label{tab:s-seed44}
\end{table}

\section{Trigger-Boundary Evolution}
\label{app:trigger}
Trigger evolution uses the same propose, evaluate, and retain-or-reject pattern as recognition evolution, but it changes only the routing description. Table~\ref{tab:s-trigger} summarizes the validation trajectories. Each run improves validation F1 while balancing recall against false-trigger rate.

\begin{table}[H]
\centering
\small
\setlength{\tabcolsep}{3pt}
\begin{tabular}{lcrrrr}
\toprule
Seed & P/R & Initial F1 & Selected F1 & Recall & FTR \\
\midrule
42 & 3/4 & .7925 & .8667 & .9028 & .1806 \\
43 & 3/5 & .8145 & .8763 & .8611 & .1042 \\
44 & 3/4 & .7820 & .8776 & .8958 & .1458 \\
\bottomrule
\end{tabular}
\caption{Trigger-description development trajectories. P/R denotes promoted/rejected proposals; FTR is the development false-trigger rate.}
\label{tab:s-trigger}
\end{table}

The held-out test comparison reported in the main paper averages the three frozen runs. Precision changes from $.831\pm.012$ to $.823\pm.031$, while recall increases from $.715\pm.042$ to $.861\pm.043$. The resulting F1 gain is 7.23 percentage points, from $.768\pm.019$ to $.841\pm.006$. Invocation success rises from $.9983\pm.0030$ to $1.0000\pm.0000$. The main effect is therefore improved coverage with a small precision cost.

\section{Human Challenge-Set Study}
\label{app:human}
\subsection{Protocol and Tie Handling}
Fifty non-expert adults aged 20--30 complete the same 40-item questionnaire. Each item asks for a minor/adult decision and a five-point confidence rating. Participants are informed that some dialogues contain psychological distress or self-harm expressions, participate voluntarily, consent to research use upon submission, and provide no direct identifiers.

Individual metrics use all 40 items. Strict-majority consensus resolves 38 items; \texttt{human-028} and \texttt{human-033} receive 25--25 votes and remain unresolved. The primary majority analysis therefore has 95\% coverage. This rule is fixed from the votes and does not use the gold labels.

\subsection{Individual Performance and Agreement}
\begin{table}[H]
\centering
\small
\begin{tabular}{lrrc}
\toprule
Individual metric & Mean & SD & 95\% CI \\
\midrule
Accuracy & .7255 & .1031 & $[.6970,.7530]$ \\
Minor F1 & .7275 & .1130 & $[.6956,.7570]$ \\
\bottomrule
\end{tabular}
\caption{Distribution of individual participant performance over all 40 challenge items. Confidence intervals use respondent bootstrap resampling.}
\label{tab:s-individual}
\end{table}

Fleiss' $\kappa$ is .4050 and mean pairwise agreement is .7034. Table~\ref{tab:s-stratum} shows that Boundary and Hard items produce the lowest accuracy. Empirical correctness increases from 55.2\% at confidence 1 to 79.0\% at confidence 5, indicating that confidence is informative but imperfect.

\begin{table}[H]
\centering
\small
\begin{tabular}{lrr}
\toprule
Stratum & Items & Human accuracy \\
\midrule
Easy & 10 & .834 \\
Hard & 15 & .657 \\
Adversarial & 10 & .808 \\
Boundary & 5 & .548 \\
\bottomrule
\end{tabular}
\caption{Mean item-level human accuracy by construction stratum.}
\label{tab:s-stratum}
\end{table}

\subsection{Human--System Comparison}
\begin{table}[H]
\centering
\small
\begin{tabular}{lrrr}
\toprule
Reference & Coverage & Accuracy & Minor F1 \\
\midrule
Mean individual & 1.00 & .7255 & .7275 \\
Strict majority & .95 & .8684 & .8571 \\
Seed 42 & 1.00 & .6000 & .5556 \\
Seed 43 & 1.00 & .7750 & .7568 \\
Seed 44 & 1.00 & .6500 & .5882 \\
\bottomrule
\end{tabular}
\caption{Human and system references on the shared challenge set. Strict-majority metrics use the 38 non-tied items; all other rows use all 40 items.}
\label{tab:s-human-system}
\end{table}

On the 38 non-tied items, strict-majority accuracy and F1 are .8684 and .8571. Seed 43 reaches .7895 accuracy and .7647 F1 on the same items. The paired F1 difference is $-.0925$ with a stratified item-bootstrap 95\% interval of $[-.2676,.0727]$; the exact McNemar test gives $p=.4531$. Under this limited item count, the best observed run is not statistically distinguishable from strict-majority consensus. Seeds 42 and 44 have lower paired accuracy, with exact McNemar $p=.0117$ and $p=.0391$, respectively.

\begin{table}[H]
\centering
\small
\setlength{\tabcolsep}{4pt}
\begin{tabular}{lrrr}
\toprule
Tie rule & Coverage & Accuracy & Minor F1 \\
\midrule
Abstain on 25--25 & .95 & .8684 & .8571 \\
Map both ties to adult & 1.00 & .8250 & .8108 \\
Map both ties to minor & 1.00 & .8750 & .8718 \\
\bottomrule
\end{tabular}
\caption{Sensitivity of strict-majority results to the two unresolved items. The primary analysis uses abstention.}
\label{tab:s-ties}
\end{table}

Both unresolved items have minor gold labels, which explains the direction of the two deterministic mappings in Table~\ref{tab:s-ties}. The abstention rule remains primary because it is independent of those labels.

\section{Frozen Artifacts and Release Scope}
\label{app:release}
The frozen experiment records include split manifests, prediction files, per-run metric reports, prompt and schema resources, candidate diffs, promotion decisions, Maker--Checker materials, questionnaire mappings, and statistical reanalysis outputs. These artifacts support the reported numbers and version histories. The authors have confirmed that the experimental code will be released after publication under a license that permits research use. Dataset release remains subject to third-party redistribution rights; independently releasable generated records, manifests, and documentation will be separated from restricted source material.

\end{document}